# Image preprocessing and modified adaptive thresholding for improving OCR

Rohan Lal Kshetry

Jadavpur University, Kolkata

**Abstract:**

In this paper I have proposed a method to find the major pixel intensity inside the text and thresholding an image accordingly to make it easier to be used for optical character recognition (OCR) models. In our method, instead of editing whole image, I are removing all other features except the text boundaries and the color filling them. In this approach, the grayscale intensity of the letters from the input image are used as one of thresholding parameters. The performance of the developed model is finally validated with input images, with and without image processing followed by OCR by PyTesseract. Based on the results obtained, it can be observed that this algorithm can be efficiently applied in the field of image processing for OCR.

**Keywords:** OCR, image processing, adaptive thresholding, segmentation, text recognition

## 1. Introduction:

A lot of information nowadays is shared in image formats, either by taking a photo from a screen, a screenshot or digitized from a paper [3]. Nowadays, Optical Character Recognition or OCR models help to retrieve the texts out of these images. But the efficiency and accuracy of OCR depends largely on the quality of the image like clarity, noise, text to background contrast etc. In the past 50 years character recognition has been studied extensively [4], this paper mainly focuses on performing OCR on photographs.

Several algorithms have been proposed to perform OCR on images with texts. Tesseract is one of such models that was initially HP Research Prototype at *UNLV 4th Annual Test of OCR Accuracy [2]*.

In Tesseract, the outlines of the components are stored first also known as connected component analysis. These outlines are gathered into blobs. These text lines are generated by gathering these blobs, which are broken into words according to character spacing. Recognition of texts is a two-pass process [1].

In the first pass, if a word is recognized satisfactorily it is passed to the training data. This helps the classifier to recognize the text more accurately as lower down the page. The second pass is used to recognize the texts on the top of the page that were not properly recognized in the first pass [1].

Optical Character Recognition (OCR) works best when there is a clear segmentation between text and the background. These segmentations need to have high dots per inches (DPI) as well. In practice the difference between text and the background varies from image to image, also the color filling the texts might not be uniform. Sometimes OCR models find it difficult to read even screenshots taken in mobile phones. This variation in types of images makes the generalization of segmentation more challenging.

The LCD or LED screens are made up of an array of red, green and blue dots; this ends up being similar in size to the red, green or blue samplers in camera. This similarity gives rise to generation of a specific pattern called 'Moire Pattern'. This adds noise to the image taken from the screen using a camera. Similarly, refresh rate also plays a role to reduce smoothness in the image.

The ability of an OCR model to read text from an image largely depends on the image quality. Once a digital image is passed through any printed form and a photo is taken from the printed image, the quality and information density are immediately reduced. In real life, this step gets repeated again and again, finally resulting in a very poor image quality. These effects make the segmentation of the text from background difficult, finally reducing the efficiency of OCR models.

In our present study, I have developed a simple algorithm to find the major pixel intensity inside the text and apply image specific adaptive thresholding to make the text appear clearer to the background.

**2. Methodology:**

**2.1. Detecting and cropping one letter**

The input image is converted into a NumPy array of dimension equal to the number of pixels on horizontal axis and vertical axis respectively. Each element of the array is an array of three elements representing the intensity of red, green and blue colors in each pixel of the input image, e.g. [[11, 2, 210], [2, 130, 115] ….].

Using PyTesseract library the coordinates of boxes bounding each detected letter are found. These coordinates are used to crop the piece of the input image containing nearly one letter from the input image.

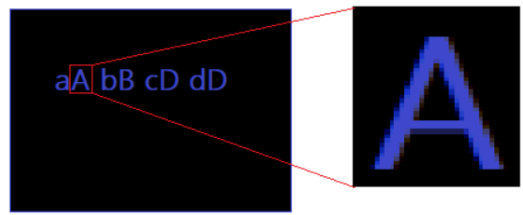

Fig 1. Cropping detected letter from image

**2.2. Converting to gray-scale and primary thresholding**

This cropped image is converted to a grayscale image, where all the pixels are converted to single channel value representing the intensity of black. This reduces the number of variables of a pixel from 3 to 1. This would not affect the performance of OCR as the gradient is more important in OCR than the color.

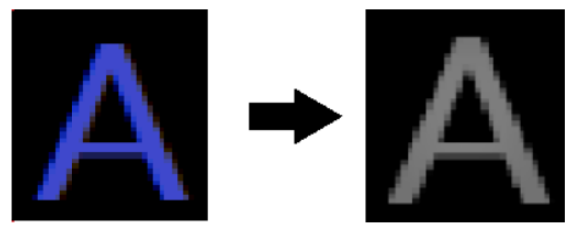

Fig 2. RGB to gray-scale conversion

The average values of each pixel inside this cropped image are calculated and used to threshold the image in such a way that if the pixel intensity value is lower than the mean value obtained above, the pixel is considered to be black, otherwise it remains unaltered. This creates a better contrast between the background and the letter in the image. It is to be noted that the threshold value is taken slightly higher than the mean value which is tuned based on previous model results. These alterations assist the model to find the outline contours of the letter.

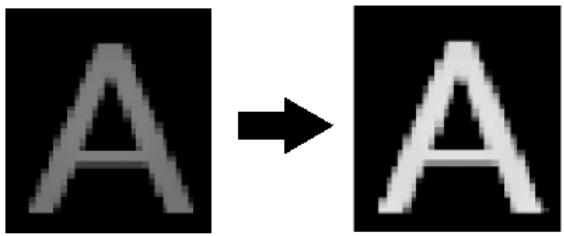

Fig 3. Primary thresholding using mean intensity

Now that the outline contour of one letter from the initial input image is obtained, starting from the top left pixel and the pointer moves to the end of the row and followed by next rows to get the coordinate and the intensity of black at that pixel. Now, if the point lies outside the contour, it is ignored. But for all the points inside the contour, the least distance of that point from the contour is calculated and saved in a list. The pixel intensity of points with highest distance from contour are taken as the thresholding value to be applied to the original image.

## 3. Results:

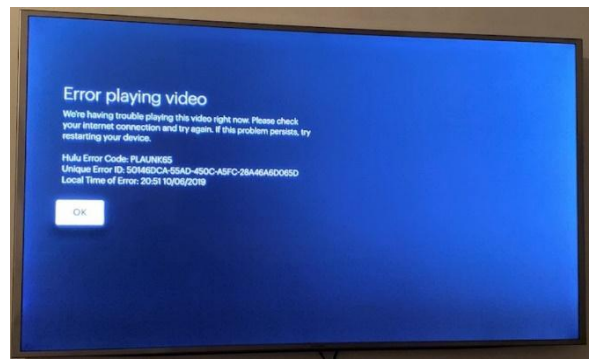

Fig 4. Input image for OCR

From the image in Fig 4, the code detected the letter 'y' and the cropped output image is shown in fig. below.

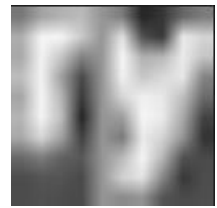

Fig 5. Cropped letter 'y'

The cropped letter is then magnified for better visibility and easier contour highlighting of the letter. From this magnified image the mean intensity is calculated; for this case it comes out to be 184.0.

The figure below (fig. 6) shows the primary thresholding step using the mean value of the intensity of all pixels. On comparing the image above (fig. 5) and the image below (fig. 6) it becomes obvious that primary thresholding makes the separation between the text and the background more prominent.

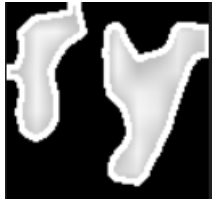

Fig 6. Primary thresholding cropped letter and highlighting outline contours

Using our algorithm, the calculated value of intensity of color filled inside the letter comes out to be 233. But as this is the pixel farthest away from the boundary, there is a possibility that this is the brightest point in the letter. So, the intensity to be used is taken as 213 instead of 233 to take the immediate darker pixels into account.

In section 3.1 the OCR result obtained by using PyTesseract directly onto the input image without any alteration to the image is shown.

**3.1. PyTesseract result on input image:**

Error playing video

We're having trouble playing this video right now Please check

'your internet connection and

'estarting your device.

try again. if this problem persists, try

Now, we will perform the image processing operation as discussed above and apply the same to the input image. Thereafter, perform optical character recognition (OCR) using PyTesseract again. This will show us the change in performance of the OCR due to applying the suggested image preprocessing and adaptive thresholding. Compared to the input image, the output image as shown below (fig. 7) the text stands out a lot brighter to the background.

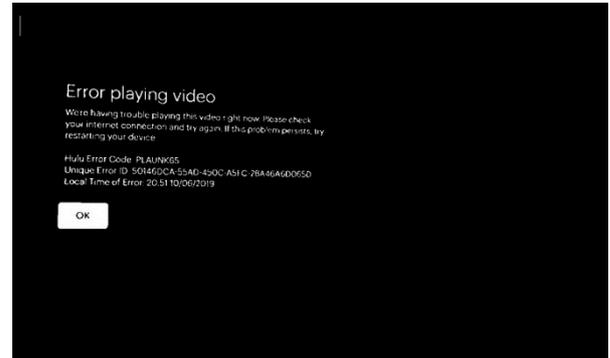

Fig 7. Final output image for OCR

**3.2. PyTesseract result on modified image:**

Error playing video

Were having trouble playing this video eght now Please check

Your imternet connection and try again. If this prob'em persists, ry

festarting your device

Hulu Error Code PLAUNK6S

Unique Error ID SO1M46DCA-59A0-450C-ASEC-28A46A500850

Local Time of Error: 20.51 10/06/2019

**4. Conclusion and future scope:**

We have represented an algorithm for image preprocessing to make the image more readable for OCR models. It can be seen that, on applying the algorithm, performance of the optical character recognition (OCR) improves drastically.

Although some of the spellings are wrongly detected, this problem can be easily solved by spelling correction models.

## 5. Acknowledgement:

I am grateful to Simplify360 India Pvt Ltd for providing with support, resources and knowledge to complete this research.

And finally, thanks to my parents who endured this long process with me, always offering support and love.